\documentclass[sigconf]{acmart}

\usepackage{newtxtext}

\usepackage{longtable}
\usepackage{tabularx}
\usepackage{colortbl}
\DeclareUnicodeCharacter{17F}{{%
		\fontencoding{TS1}%
		\fontfamily{lmr}%
		\selectfont s%
}}
\usepackage{microtype}

\AtBeginDocument{%
  \providecommand\BibTeX{{%
    \normalfont B\kern-0.5em{\scshape i\kern-0.25em b}\kern-0.8em\TeX}}}

\setcopyright{None}
\copyrightyear{2020}
\acmYear{2020}
\acmDOI{}

\acmConference[DTUC '20]{Proceedings of the 2nd International Digital Tools \& Uses Congress}{October 15--17, 2020}{Online, Tunisia}
\acmBooktitle{Proceedings of the 2nd International Digital Tools \& Uses Congress (DTUC '20), October 15--17, 2020, Online, Tunisia}
\acmISBN{}
\acmDOI{10.1145/3423603.3423996}

\begin{document}

\title{Standardizing linguistic data: method and tools for annotating (pre-orthographic) French}

\author{Simon Gabay}
\email{forename.name@unine.ch}
\affiliation{%
  \institution{Universités de Neuchâtel et de Genève}
  \city{Neuchâtel and Genève}
  \state{Switzerland}
}

\author{Thibault Clérice}
\author{Jean-Baptiste Camps}
\email{forename.name@chartes.psl.eu}
\affiliation{%
  \institution{École des Chartes}
  \city{Paris}
  \state{France}
}

\author{Jean-Baptiste Tanguy}
\email{forename.name@sorbonne-universite.fr}
\affiliation{%
  \institution{Sorbonne Université}
  \city{Paris}
  \state{France}
}

\author{Matthias Gille-Levenson}
\email{forename.name@ens-lyon.fr.}
\affiliation{%
  \institution{École normale supérieure de Lyon}
  \city{Lyon}
  \state{France}
}

\renewcommand{\shortauthors}{Gabay et al.}

\begin{abstract}
  With the development of big corpora of various periods, it becomes crucial to standardise linguistic annotation (\textit{e.g.} lemmas, POS tags, morphological annotation) to increase the interoperability of the data produced, despite diachronic variations. In the present paper, we describe both methodologically (by proposing annotation principles) and technically (by creating the required training data and the relevant models) the production of a linguistic tagger for (early) modern French (16-18th\,c.), taking as much as possible into account already existing standards for contemporary and, especially, medieval French.
\end{abstract}



\maketitle

\section{Introduction}
If medievalists have been working for years on the creation of high quality corpora as well as the tools to build and analyse them (\textit{e.g.} the \textit{Base de français médiéval} (BFM)~\cite{guillot_base_2017} and TXM~\cite{heiden_txm_2010}), the situation is different for more recent periods of French literature. As far as we know, for texts written between the 16th and the 18th\,c., \textit{Presto} is the only project that has developed a corpus~\cite{blumenthal_presto_2017} with some computational resources~\cite{diwersy_ressources_2017}. However, both are not maintained anymore, which leaves more than three centuries of French literature computationally understudied. It becomes therefore important to overcome this difficulty, and to create the tools that we desperately need for (early) modern French (henceforth EMF).

Such a task cannot be done without taking into account longstanding annotation practices, in order to allow (minimal) interoperability with already existing datasets. uch a statement is sadly easier said than done, because EMF is an intermediary stage between medieval (12th-15th\,c.) and  late modern and  contemporary (from c.\,1750) French, two states of language that tend to have different needs regarding annotation: EMF is then caught in between two (potentially incompatible) practices, one for each extreme of the continuum. Facing such a dilemma, we have decided first to look up, in diachrony, and align (when possible) our choices with those of medievalists, rather than to look immediately down at what synchronists do – which tends to be the norm. Our project is indeed to create a tagger for EMF producing data as much as possible compatible with existing medieval corpora, and therefore build the technical frame for the computational study of pre-orthographic French, understood as the forms of this tongue prior to their standardisation based on the recommendations of the \textit{Académie française}.

Textual production before the implementation of this "academic" French (c.\,mid-18th\,c., rather than the publication of its \textit{Dictionnaire} in 1694) share common features~\cite{catach_histoire_2001}. Among many others, the most important one is to undergo a significant graphematic variation, \textit{i.e.} a relative instability regarding the spelling (\textit{estoit} vs \textit{étoit}) and the segmentation (\textit{à fin que} vs \textit{afin que}). Thus, to bypass this problem, the linguistic annotation has to be thought as a pivot format~\cite[p. 393]{sylvain_les_2000} that allows users to navigate across heterogeneous data.

Several options are available, such as the conflation by phonetic identity or by lemma~\cite{Jurish2012} – proto-forms being currently investigated as a potential solution, without being yet fully operational~\cite{fourrier2020}. Among all possibilities, we have decided to opt for lemmatisation+POS tagging as a joint solution for two main reasons: on the one hand many resources are already available (cf.\,\textit{infra}), and on the other hand it is already used for other tasks such as collation~\cite{camps2019} or stylometric analysis~\cite{camps_why_2019}.

\section{Annotation principles}
Since all choices have been carefully discussed in our annotation manual~\cite{gabay2020} we will here present the main problems raised by annotating EMF: tokenisation, lemmatisation, POS and morphological tagging.

\subsection{Tokenisation}
As any language, French has undergone phenomena of agglutination at the graphematic (\textit{ce pendant} $\to$ \textit{cependant}) and at the lexical (\textit{e.g.} \textit{bien que}) level. Such a process is not easily datable, because it varies from one region or one phenomenon to the other: in long diachrony, it is therefore impossible to decide precisely when such occurrences have to be tagged as two, or one token. For this reason, we have decided to use the blank space as the main separator for the tokenisation, with hardly any exception to this rule: all locutions, may they be conjunctive (\textit{tandis que}), adverbial (\textit{à demi}) or prepositional (\textit{quant à}), are segmented in two parts rather than one (\texttt{tandis} and \texttt{que}, \texttt{à} and \texttt{demi}, \texttt{quant} and \texttt{à}). 

Following this maximalist approach, none of the compounds are analysed as one token: \textit{peut être}, \textit{long temps},\dots are all considered as multiple tokens (\texttt{peut} + \texttt{être}, \texttt{long} + \texttt{temps}) – and this despite the presence of the hyphen, which is considered as a token itself (\textit{peut-être} $\to$ \texttt{peut} + \texttt{-} + \texttt{être}, \textit{long-temps}  $\to$ \texttt{long} + \texttt{-} + \texttt{temps}\dots). For the same reason (absence of blank space), words that are already welded (\textit{monsieur} $<$ \textit{mon sieur}) are treated as one token.

Such a choice to respect the actual segmentation of original witnesses introduces a bias, because it follows the \textit{scripta} (\textit{i.e.} the writing practice) of printers in (early) modern France~\cite{pellat_levolution_1995} and not necessarily the preference of authors. This decision is however more in line with our philological approach, because it relies less on today's linguistic understanding of EMF and outsources complex decisions to historical speakers.

\subsection{Lemmatisation}
This tokenising strategy creates some lemmatisation problems. The most important one is the absence of coherent lemmas for agglutinated forms that have no autonomy outside a locution such as \textit{parce} (in \textit{parce que}, $<$ \textit{par ce}) or \textit{afin} (in \textit{afin que}, $<$ \textit{à fin}), but also with portmanteau words (\textit{tresobeissant} $<$ \textit{tres}+\textit{obeissant}). A simple solution is the creation of compounded lemmas, with an underscore separating the original lemmas (\texttt{tres\_obeissant}), but such a method cannot be generalised: on the one hand, in diachrony, (too) many tokens could need one (\texttt{mon\_sieur}? \texttt{autre\_fois}?), and on the other hand we take the risk to end up with (too) many compounded lemmas for a single token (\textit{audit}$\to$\texttt{à\_le\_dit}). In order to limit our logic, tokens that have subsisted into contemporary French are therefore analysed as a simple lemma (\texttt{parce}, \texttt{afin}), but those that have not subsisted are analysed as a compounded lemma (\texttt{tres\_obeissant}):  this limits the creation of lemma, and maintains interoperability with other language states.

For other words, we have decided to use, when possible, the contemporary form of each token as a canonical form. It usually is the infinitive form (for verbs: \textit{mangeons} $\to$ \texttt{manger}) or the masculine singular (for nouns or adjectives: \textit{comtesses} $\to$ \texttt{comte}). Such a process is diachronically problematic, since some words have lost  (\textit{dominus}/\textit{domina} $>$ \textit{dame} but also old fr. \textit{dom}), or almost lost (\textit{damoiseau} vs \textit{demoiselle}), their masculine counterpart over time: in such cases, we only choose the feminine version if the masculine one is not attested anymore at the end of the 18th\,c.

In order to control the consistency of the annotation, several authority lists have been created. The main one derives from the lexicon of medieval French inflected forms \textit{LGeRM}~\cite{souvay_lgerm_2009} in its \textit{mode} version (\textit{i.e.} \textit{moderne étendu}) produced for the \textit{Presto} project~\cite{diwersy_ressources_2017}. The \textit{LGeRM} lexicon is based on the \textit{Freeling} version of the Le\textit{fff} ~\cite{sagot_lefff_2010}, augmented with \textit{Morphalou}~\cite{romary_standards_2004} for contemporary lemmas, and the \textit{Dictionnaire du Moyen Français} (DMF)~\cite{DMF} for (early) modern ones. Such an approach should help us maintain minimal interoperability with the medieval version of \textit{LGeRM}, but also the \textit{Trésor de la Langue Française informatisé} (\textsc{TLFi})~\cite{TLFi}. Two other authority lists have been created: one for named entities (people, places\dots) and an other one with foreign words.

\subsection{POS-tagging}
Regarding POS-tagging, we have decided to use CATTEX~\cite{prevost_jeu_2013} for three reasons: first it is already used for the \textit{BFM} and by several other corpora~\cite{deucalionFro}, second it already benefits from a detailed annotation manual~\cite{guillot_manuel_2013}, third, it has been designed to cope with the complexity of old states of the French language.

Between the options exposed by~\citeN{guillot_principes_2013}, we have opted for morpho-syntactical annotation: we take into account the context when categorising tokens and adjectives can be tagged as adverbs (\textit{il parle fort}) or adverbs as substantives (\textit{il fait le bien}). It allows us to follow the evolution of uses over time for a single word. However, following CATTEX as well as \textit{Presto} recommendations, we never categorise present and past participles as adjectives, because if such distinctions are already difficult in synchrony, they are even harder to identify in diachrony. It would be too complicated to precisely differentiate the past participle of \textit{perdre} (\textit{perdu}) from the adjective \textit{perdu} over the very centuries during which the verb has undergone a slow process of adjectivisation, especially without clear markers such as the determiner for the substantivization (\textit{il fait bien} vs \textit{il fait le bien}).

Our (maximalist) tokenisation principles incitate (but do not force) us to tag words with their original part of speech: \textit{bien} in \textit{bien que} is an \texttt{ADVgen} and not a member of a potential \texttt{LOCconj} (that does not exist within CATTEX). This choice considerably simplifies the interoperability with Old and Middle French. It does create some problems though, and  we are forced to use analogies to tag certain tokens: \textit{pource} in \textit{pource que} is therefore \texttt{ADVgen}, such as \textit{bien} in \textit{bien que}. 

\subsection{Morphology}
We have decided to implement CATTEX-max~\cite{prevost_jeu_2013}, that is to say to annotate each token with morphological information (gender, mood, tense\dots). Contrary to lemmas and POS, this linguistic information has been added for technical purposes (it has helped maximise the efficiency of the final model): if it has been proofread to avoid major mistakes, its perfection is not guaranteed.

\section{Training a model}
Based on the previously introduced annotation choices, we have decided to train a lemmatiser and a POS-tagger.

\subsection{Data recycling}

A substantial amount of heterogeneous data is already available (cf.\,tab.~\ref{tab:corpus}). Considering the cost of creating a gold corpus, we have decided that it would be more efficient to re-process existing data rather than create new ones from scratch.

    \begin{small}
        \begin{table}[h]
            \centering
            \begin{tabular}{llllll}
            \toprule
                Name          & Gold & Norm. & Tokens    & POS     & Morph \\
                \midrule
                \textit{CornMol}       & Yes  & Yes   &    90,000 & CATTEX  & Yes   \\ 
                \textit{FranText OA}   & No   & Yes   & 2 400,000 & EAGLES   & No    \\   
                \textit{Presto gold}  & Yes  & Yes   &    60,000 & MULTEXT & No    \\   
                \textit{Presto core} & No   & Yes   & 6,820,000 & MULTEXT & No    \\  
                \textit{Presto controlled} & No   & Yes   & 11,636,000 & MULTEXT & No    \\  
                \textit{Presto extended} & No   & Yes   & 28,309,000 & MULTEXT & No    \\  
                \bottomrule
            \end{tabular}
            \caption{Available training data}
            \label{tab:corpus}
        \end{table}
    \end{small}
    
    Two corpora come from a previous study~\cite{Camps2020} and are almost ready to use:
    \begin{itemize}
        \item The \textit{CornMol} corpus has been created to carry stylometric studies~\cite{camps_why_2019}. It is based on 41 comedies written in the 17th\,c., carefully sampled and proofread, which have been thoroughly described~\cite{Camps2020}.
        \item The \textit{Frantext open access data}~\cite{frantext} is composed of 32 texts, mainly written in the 18th (7 texts), 19th (24 texts) and 20th\,c. (4 texts). Because the corpus is already tagged and lemmatised (but not fully corrected) following other guidelines than ours, the lemmas have been aligned according to our standards, as much as we could.
    \end{itemize}
    
    Regarding \textit{Presto}, because it is composed of four different subcorpora (for a detailed description of the different levels and variations, cf\,~\citet{Vigier2018}), it requires some selection, processing and correction:
    
\begin{itemize}
    \item A gold corpus is made of 60 000 tokens, taken from 5 texts written in the 16th (1 text), 17th (2 texts) and 18th\,c. (2 texts). These texts have all been sampled, tokenised, tagged, lemmatised and proofread to create training data for a \textit{TreeTagger} model.
    \item The final \textit{Presto} corpus is a three-fold one: \textit{noyau} (``core''), \textit{contrôlé} (``controlled'') and \textit{étendu} (``extended''). We have limited ourselves to a selection of its core version (cf.\,Appendix
    ).
\end{itemize}

\subsection{Preparation}

Because the \textit{CornMol} and \textit{Frantext Open Access} corpora have already been prepared for our previous experiment on normalised-spelling French, it already follows closely our choices regarding tokenisation and POS-tagging (minor changes have been made in between), and the shift of authority list from \textit{Morphalou} to \textit{LGeRM} is of almost no impact since the latter derives from the former. All the effort have therefore been put on the correction and the alignment on our standards of the two \textit{Presto} sub-corpora, which was problematic for three reasons. First, our choices differ strongly from \textit{Presto}'s regarding tokenisation and POS-tagging – but not lemmatisation, since they also use \textit{LGeRM}. Second, because their training data has been mainly prepared for POS-tagging and not lemmatisation, and also because the annotation has been done with an older and less precise tool (\textit{TreeTagger}), the result is far from being perfect. Third, it is impossible to recycle some texts of their final corpus into training data, because we need (fairly) clean texts – Jean de Léry's \textit{Voyage en terre du Brésil} had therefore to be withdrawn because of the noise produced by hyphenation. Most of the correction work has been done with \textit{Pyrrha}~\cite{pyrrha_2019}.

In order to deal with more complex data than the one we use for training, we have artificially added a very limited number of glyphs typical of (early-)modern prints such as the long \textit{s} (\textit{ſ}), the  \textit{eszett} (\textit{ß}) or tilded letters traditionally used as abbreviations (\textit{õ}, \textit{ã}, \textit{ũ}, \textit{ĩ}).

Out of all these data, two different corpora have been produced:
\begin{itemize}
    \item A primary corpus, fully annotated with lemma, POS and morphology, based on tokens with normalised and non-normalised spelling, has been created out of the two gold corpora (\textit{CornMol} and the revised version of \textit{Presto} gold).
    \item A secondary corpus, with lemmas only, has been produced out of \textit{Frantext open access data}, the corrected versions of \textit{Presto} gold and \textit{Presto} core, and \textit{CornMol}.
\end{itemize}

These two corpora are used to train two separate models: one for the POS and the morphology, and another for lemmatisation only. Therefore, two training datasets have been created using \textit{Protogénie}~\cite{protogenie} (cf.\,tab.~\ref{tab:training_Sets}): they contain a train set, a development set (for evaluation during training) and a test set (for in-domain testing). The breakdown of tokens between the train, dev and test set is dependent on the the total amount of data and has been done on an empirical basis, in order to keep as much tokens for training as possible, while still keeping a reliable test set. As such, it is dependent on the total size of the corpus (the bigger the corpus, the smaller the fraction of it which is necessary for reliable training).

\begin{table}[H]
    \centering
    \begin{tabularx}{\columnwidth}{XXXXX}
    \toprule
        Corpus    & Train set & Dev set  & Test set & Total  \\
    \midrule
        \textit{POS}       & 132,905   & 9,733    & 15,303   & 157,941   \\
                  & 84\%      & 6\%      & 10\%     &        \\
    \midrule
        \textit{Lemma}     & 6,666,473 & 70,009   & 352.482   & 7,088,964 \\
                  & 94\%      & 1\%      & 5\%      &         \\
    \bottomrule
    \end{tabularx}
    \caption{Breakdown of the tokens per set (numbers are rounded)}
    \label{tab:training_Sets}
\end{table}

    Because of the size of the corpus, it is impossible to control each token. The general quality is therefore guaranteed by the use of already existing corpora, which have already been carefully prepared for previous tasks. The consistency of the annotation is controlled with authority lists, but also with intermediary models, which help inspect the consistency of the data sets. After each training, the creation of a confusion matrix allows to identify frequent confusions between predictions and ground truth data, that could be due to inconsistencies in the human annotation.
    
\subsection{Set up}

We use \textit{Pie}~\cite{pie_2019} as our main training and tagging software\footnote{Previous tests included a comparison with \textit{Marmot} for POS-tagging, and results were equivalent to those obtained with \textit{Pie}~\cite{Camps2020}}. For training, \textit{Pie} is used for lemma- and POS-tagging. At inference time, we use \textit{Pie-extended}~\cite{pieExtended} which wraps the former with more user-friendly functionalities (model sharing, advanced post- and preprocessing options\dots).
Several architectures have been tested for our previous study on normalised texts (for more detailed scores, cf.~\citet{Camps2020})\footnote{``Unknown tokens'' are tokens never seen during training, while ``ambiguous tokens'' are forms that can correspond to different lemmas. ``Unknown targets'' are lemmas never seen in training, but that the neural network can still sometimes accurately predict, thanks to its character level modelling.}, the best one being:
\begin{itemize}
    \item RNN with 300 dimensions for characters
    \item Sentence and word embeddings (150 dimensions) using 343 drama texts from \cite{fievre_theatre_2007} and those of the \textit{FranText Open Access} that we presented \textit{supra}, for a total of c.\,7M tokens.
    \item Hidden size of 150
    \item Forward and backward language models
\end{itemize}

However, additional tests have proven since the possibility of increasing even more the accuracy with a bigger number of dimensions for the RNN (300$\to$400) and the hidden size (150$\to$256), and NFKD unicode normalisation of the data, without word embeddings.

\begin{table}[htbp]
    \centering \small %
    \begin{tabularx}{\columnwidth}{l|XXXXX}
    \toprule
        \multicolumn{6}{c}{LEMMA} \\
    \midrule
        \textit{Lemma}         & M2 & M3 &  \multicolumn{2}{c}{M3+NFKD}  & \textit{support} \\
    \midrule
        \textit{In-domain}     & 99.09 & 99.12    & \multicolumn{2}{c}{\textbf{99.28}} & \textit{4,181} \\
        \textit{Out-of-domain} & 97.92 & 98.59    & \multicolumn{2}{c}{\textbf{98.67}} & \textit{13,497}\\
    \midrule
        \multicolumn{6}{c}{POS} \\
    \midrule
        POS & M1  & M1+aux           & M2 & M2+aux           & \textit{support}\\
    \midrule
        \textit{All}             & 96.72          & 96.51          & 96.84          & \textbf{97.01} & \textit{4,181}\\ 
        \textit{Ambiguous token} & 91.86          & 91.43          & \textbf{92.40} & 92.29          & \textit{934}\\ 
        \textit{Unknown tokens}  & \textbf{86.24} & \textbf{86.24} & 78.44          & 81.65          & \textit{218} \\
    \bottomrule
    \end{tabularx}
    \end{table}
\begin{table}[htbp]
    \centering \small %
    \caption{Scores (normalised spelling data only). {\normalfont \textit{M1} is the basic \textit{Pie} configuration, \textit{M2} adds word embeddings to \textit{M1} architecture, and \textit{+aux} indicates the use of auxiliary tasks – these architectures being fully described in~\citet{Camps2020} as \textit{base (sent-lm)} and \textit{wembs}. \textit{M3} uses a bigger number of dimensions (400) for the RNN than for the \textit{M1} configuration (150), and \textit{+NFKD} indicates the additional use of unicode normalisation.}}
    \label{tab:pie_accs_norm}
\end{table}

Based on these new results, we have decided to test this new successful \textit{M3} configuration rather than the \textit{M1} and \textit{M2} previously presented as the most efficient. Two versions have been tested: the first with one linear layer, the second with two linear layers. For each configuration, due to the stochastic nature of the process, three models were trained, using early stopping with threshold 0.001 and patience 7. The best one was retained. For time and ecological purposes, NFKD normalisation has been tested on the most relevant architecture only.

\subsection{Results}

First, the existence of spelling variation seems to have a limited impact, or to be efficiently counter-balanced by our data augmentation: the accuracy remains similar to the one of models trained exclusively on normalised spelling data for lemmatisation (98.62\% vs 99.28\%) and even higher for POS-tagging (97.18\% vs 97.01\%). Such results are very promising. When looking into the details, we observe that two linear layers have an almost null (+0.001 pt of \% for POS) or slightly negative impact (-0.0002 pt of \% for lemmas). Considering the (much) larger training time and energy use, it does not seem to be a satisfactory solution. 

\begin{table}[htbp]
    \centering \small %
    \begin{tabularx}{\columnwidth}{l|XXXX}
        \toprule
                          & 1 layer         & +NFKD   & 2 layer  & \textit{support} \\
        \midrule
\textit{all}              & \textbf{98.62} & 98.48  & 98.46   & \textit{352,483}  \\
\textit{unknown tokens}   & \textbf{68.96} & 68.07  & 68.31   & \textit{3,843}  \\
\textit{ambiguous tokens} & \textbf{98.3}  & 98.16  & 98.12   & \textit{182,472}  \\
\textit{unknown targets}  & \textbf{54.27} & 54.13  & 54.32   & \textit{2,047}  \\
        \bottomrule
    \end{tabularx}
    \caption{Lemmas (normalised+original spelling)}
    \label{tab:pie_accs_lemmas}
\end{table}

Because NFKD lowers the amount of characters (235 $\to$ 164) we have decided to lower the amount of dimensions for the RNN character embeddings (400 $\to$ 300). The final results show a slightly negative impact on the accuracy, which proves that the efficiency of NFKD on the \textit{CornMol} corpus is linked to the smaller size of the corpus and that unicode decomposition and normalisation is probably less relevant on big corpora – similar scores could be achieved with fine tuning, especially by lowering the number of dimensions.

\begin{table}[htbp]
    \centering
    \begin{tabularx}{\columnwidth}{l|XXX}
        \toprule
                          & 1 layer & 2 layers & \textit{support} \\
        \midrule
\textit{all}              & 97.09  & \textbf{97.18}   & \textit{14,303}  \\
\textit{unknown tokens}   & 87.57  & \textbf{88.4}    & \textit{724}  \\
\textit{ambiguous tokens} & 93.81  & \textbf{94.11}   & \textit{4,281}  \\
        \bottomrule
    \end{tabularx}
    \caption{POS (normalised+original spelling)}
    \label{tab:pie_accs_pos}
\end{table}

The significant increase of training data (90,000 to 150,000 tokens) have a significant effect on the efficiency of the POS-tagging model, and we improve the results of the previous model based on normalised-spelling data only.

\subsection{Impact of spelling variation}

The score for ``ambiguous tokens'' given by \textit{Pie} evaluates its efficiency regarding homographs, \textit{i.e.} its capacity to disentangle tokens like \textit{entre} which can be both a form of the verb \textit{entrer} (\textit{il entre dans la pièce}) or the preposition (\textit{entre les murs}). However, lemmatising historical variation-rich languages is also about handling polymorphism, because a same word can frequently have different graphematic realisations (\textit{e.g.} \textit{besoin} vs \textit{besoing}). We have therefore decided to evaluate the robustness of our best model regarding the graphematic volatility of (early-modern) texts, as \citet{Millour2019a} did for their study on dialectal variation.

214 pairs of two different tokens sharing the same lemma, POS-tag and morphology have been extracted from the whole \textit{presto gold} corpus, so that we have a \textit{Form A} and a \textit{Form B} in each pair: \textit{afin} vs \textit{affin}, \textit{changements} vs \textit{changemens}\dots. 
It is to be noted that, by including only pairs that are actually present in the final training data, we have limited the significance of our results for out-of-domain and/or rare graphematic realisations.

In order to evaluate the effect of spelling variation, we designed a simple algorithm: in the test set, for each pair, all sentences containing a token matching the lemma and one of the variant of the pairs are extracted.
This group of sentences is then duplicated: the first set is left untouched, while, in the second set, all occurrences of variant $i$ are replaced by the alternative variant $j$. Both groups are then lemmatised and the accuracy is compared to evaluate the robustness of the model despite spelling variation in  strictly equivalent contexts.

By doing so, we are able to evaluate the recognition of the spelling variants while neutralising variations due to the left and right contexts (each form is artificially given the same number of occurrences and the same contexts).
We then test our model on both set, and compare accuracy for $i$ ($Acc_i$) and for $j$ ($Acc_j)$, with a simple difference:
\[
\Delta_{Acc} = | Acc_i - Acc_j |
\]

For all these pairs, the median $\Delta_{Acc}$, the geometric mean $\Delta_{Acc}$, and the weighted geometric mean, using the frequency of variant $j$ in the training data as weight, are all equal to 0  (arithm. mean is 0.025; weight. arithm. mean is 0.011).
We can therefore conclude that spelling variation has a very limited impact on the accuracy, as long as both forms are present in the training data. In the future, it would be necessary to extend this kind of analysis to pairs for which one variant is unseen by the model\footnote{%
   The scripts for this particular evaluation are available on: Thibault Clérice and Jean-Baptiste Camps, \texttt{ PonteIneptique/classique\_variante}, \textit{Github}, 8 sept. 2020, \url{https://github.com/PonteIneptique/classique\_variante}.
}.

\subsection{Out-of-domain tests}

We have tested our best lemma and our best POS models on the same out-of-domain testing data than the one used for our previous study. There are two test sets for each century: one made only of theatre, the other one of everything but theatre. Each test set is composed of 10 short samples (c.\,100 tokens), as representative as possible of the linguistic production of the century (female and male authors, decade of publication, genre\dots).

Because the model built for the previous study was trained on normalised-spelling data, out-of-domain data had been normalised too. For this new study, the out-of-domain test set has been duplicated and now exists in both original and normalised transcriptions for the 16th, 17th and 18th\,c. texts only, and not for 19th and 20th\,c. texts, which do not require any spelling normalisation.

\begin{table}[htbp]
    \centering
    \begin{tabular}{c|ccccc|c}
    \toprule
        Corpus & \textbf{16th} & \textbf{17th} & \textbf{18th} & \textbf{19th} & \textbf{20th} & \textit{All cent.} \\
    \midrule
    \multicolumn{7}{c}{Test 1} \\
    \midrule
        \textit{Drama}      & \cellcolor[gray]{0.8}97.6  & \cellcolor[gray]{0.7}98.10 & \cellcolor[gray]{0.7}98.88 & \cellcolor[gray]{0.7}98.34 & \cellcolor[gray]{0.7}98.00 & \cellcolor[gray]{0.7}98.19 \\
        \textit{Not drama}  & \cellcolor[gray]{0.8}97.78 & \cellcolor[gray]{0.7}98.02 & \cellcolor[gray]{0.7}98.06 & \cellcolor[gray]{0.95}96.97 & \cellcolor[gray]{0.8}97.39 & \cellcolor[gray]{0.8}97.65 \\
        \textit{Both}       & \cellcolor[gray]{0.8}97.69 & \cellcolor[gray]{0.7}98.06 & \cellcolor[gray]{0.7}98.46 & \cellcolor[gray]{0.8}97.66 & \cellcolor[gray]{0.8}97.70 & \cellcolor[gray]{0.8}97.92 \\
    \midrule
    \multicolumn{7}{c}{Test 2} \\
    \midrule
        \textit{Drama}      & \cellcolor[gray]{0.95}96.65 & \cellcolor[gray]{0.8}97.42 & \cellcolor[gray]{0.8}97.69 & \cellcolor[gray]{0.7}98.2  & \cellcolor[gray]{0.8}97.5  & \cellcolor[gray]{0.8}97.51\\
        \textit{Not drama}  & \cellcolor[gray]{0.8}97.48 & \cellcolor[gray]{0.7}98.24 & \cellcolor[gray]{0.7}98.27 & \cellcolor[gray]{0.8}97.12 & \cellcolor[gray]{0.95}96.79 & \cellcolor[gray]{0.8}97.59\\
        \textit{Both}       & \cellcolor[gray]{0.8}97.08 & \cellcolor[gray]{0.8}97.83 & \cellcolor[gray]{0.8}97.99 & \cellcolor[gray]{0.8}97.66 & \cellcolor[gray]{0.8}97.15 & \cellcolor[gray]{0.8}97.55\\
    \midrule
    \multicolumn{7}{c}{Test 3} \\  
    \midrule
        \textit{Drama}      & 93.93 & \cellcolor[gray]{0.95}96.36 & \cellcolor[gray]{0.95}96.95 & \cellcolor[gray]{0.7}98.2  & \cellcolor[gray]{0.8}97.5  & \cellcolor[gray]{0.95}96.64\\
        \textit{Not drama}  & \cellcolor[gray]{0.95}96.23 & \cellcolor[gray]{0.8}97.25 & \cellcolor[gray]{0.7}98.2  & \cellcolor[gray]{0.8}97.12 & \cellcolor[gray]{0.95}96.79 & \cellcolor[gray]{0.8}97.13\\
        \textit{Both}       & \cellcolor[gray]{0.95}95.12 & \cellcolor[gray]{0.95}96.8  & \cellcolor[gray]{0.8}97.59 & \cellcolor[gray]{0.8}97.66 & \cellcolor[gray]{0.8}97.15 & \cellcolor[gray]{0.95}96.89\\
    \bottomrule
    \end{tabular}
    \caption{Lemmatisation accuracies of the best model on out-of-domain data. {\normalfont \textit{Test 1} tests our best model trained on normalised-spelling data on normalised out-of-domain data. \textit{Test 2} tests our best model trained on original-spelling data on normalised out-of-domain data. \textit{Test 3} tests our best model trained on original data on original out-of-domain original data.}}
    \label{tab:outofdomain_lemma}
\end{table}

We observe that accuracies on normalised-spelling out-of-domain testing data are relatively similar for both the model trained on normalised data and the one trained on original data. The fact that the main difference are observed for theatrical plays could imply that the model trained on normalised data has been tailored for this genre. Without surprise, regarding non-normalised out-of-domain testing data, the accuracy of the model diminishes with centuries as we go back in time.

\begin{table}[htbp]
    \centering
    \begin{tabular}{c|ccccc|c}
    \toprule
        Corpus & \textbf{16th} & \textbf{17th} & \textbf{18th} & \textbf{19th} & \textbf{20th} & \textit{All cent.} \\
    \midrule
    \multicolumn{7}{c}{Test 1} \\
    \midrule
        \textit{Drama} & \cellcolor[gray]{0.8}95.05&\cellcolor[gray]{0.7}96.59&\cellcolor[gray]{0.8}95.98&\cellcolor[gray]{0.8}94.81&\cellcolor[gray]{0.9}93.57&\cellcolor[gray]{0.8}95.18\\
        \textit{Not drama} &\cellcolor[gray]{0.95}92.89&\cellcolor[gray]{0.8}94.27&\cellcolor[gray]{0.7}96.53&\cellcolor[gray]{0.95}91.87&\cellcolor[gray]{0.95}91.35&\cellcolor[gray]{0.9}93.42\\
        \textit{Both} & \cellcolor[gray]{0.9}93.93&\cellcolor[gray]{0.8}95.44&\cellcolor[gray]{0.7}96.27&\cellcolor[gray]{0.9}93.36&\cellcolor[gray]{0.95}92.48&\cellcolor[gray]{0.8}94.30\\
    \midrule
    \multicolumn{7}{c}{Test 2} \\
    \midrule
        \textit{Drama} &  \cellcolor[gray]{0.9}93.53&\cellcolor[gray]{0.8}95.68&\cellcolor[gray]{0.8}95.01&\cellcolor[gray]{0.9}93.37&\cellcolor[gray]{0.9}93.42&\cellcolor[gray]{0.8}94.19\\
        \textit{Not drama} & \cellcolor[gray]{0.95}92.75&\cellcolor[gray]{0.8}94.5&\cellcolor[gray]{0.8}95.21&\cellcolor[gray]{0.95}92.09&\cellcolor[gray]{0.9}93.59&\cellcolor[gray]{0.9}93.64\\
        \textit{Both} & \cellcolor[gray]{0.9}93.12&\cellcolor[gray]{0.8}95.09&\cellcolor[gray]{0.8}95.12&\cellcolor[gray]{0.95}92.74&\cellcolor[gray]{0.9}93.5&\cellcolor[gray]{0.9}93.92\\
    \midrule
    \multicolumn{7}{c}{Test 3} \\  
    \midrule
        \textit{Drama} &  89.3&\cellcolor[gray]{0.95}92.72&\cellcolor[gray]{0.8}94.49&\cellcolor[gray]{0.9}93.37&\cellcolor[gray]{0.9}93.49&\cellcolor[gray]{0.95}92.72\\
        \textit{Not drama} & 89.42&\cellcolor[gray]{0.95}92.6&\cellcolor[gray]{0.8}95.49&\cellcolor[gray]{0.95}92.09&\cellcolor[gray]{0.9}93.59&\cellcolor[gray]{0.95}92.67\\
        \textit{Both} & 89.36&\cellcolor[gray]{0.95}92.66&\cellcolor[gray]{0.8}95.01&\cellcolor[gray]{0.95}92.74&\cellcolor[gray]{0.9}93.5&\cellcolor[gray]{0.95}92.69\\
    \bottomrule
    \end{tabular}
    \caption{POS accuracies of the best model on out-of-domain data}
    \label{tab:outofdomain_pos}
\end{table}

Results regarding POS annotation are relatively similar to the one of lemmatisation, and we observe again a significant impact of the theatrical genre on the accuracy.


%

\section{Interpretation}

Interestingly, none of the most important errors are linked to graphematic polymorphism. Most of the problems are linked to homographs: \textit{le} pronoun (\textit{il le veut}) vs determiner (\textit{le château}), the verb \textit{a} (\textit{il a}) vs the preposition (\textit{à Genève}) when it does not have the grave accent (\textit{a Genève}), the determiner \textit{des} (\textit{il fait des ronds}) vs the enclise \textit{de}+\textit{le} (\textit{le père des enfants}). If such results confirm those obtained on normalised data, the presence of noise in the \textit{Presto core} corpus could explain some errors.

\section{Further Work}
The next step, already in preparation, is the extension of the annotation to named entities to train a NER for EMF. With texts published on three centuries, and especially with a long excerpt of the \textit{Encyclopédie} (c.\,1,500,000 tokens), we do believe that our corpus is rich enough to carry preliminary tests. The presence of already corrected lemmas and POS tags for proper nouns (to which a special attention was given during the correction of the corpus) should dramatically ease the annotation process. 

Regarding lematisation and POS-tagging, the most important task will be the extension of the training data for 16th\,c., and even 15th\,c. French, which is clearly underrepresented in our corpora: the complexity of middle French would certainly require more data, and new strategies to maintain the interoperability between medieval and (early) modern datasets.

\section{Contributions}
This project is a follow-up to another one designed by Florian Cafiero, J.-B. C. and S.G. It has been led by S.G., who prepared the data and the final article, with the help of many: J.-B. T. for the conversion of \textit{Presto gold} from MULTEX to CATTEX, Lucence Ing for the annotation of \textit{Presto gold}, and Matthias Gille Levenson for the correction of \textit{Presto core}. Trainings have been configured and executed by T.C. 
J.-B. C. has accompanied the entire process, providing technical guidance and philological feedback on the compatibility with medieval data.  All authors discussed the results and contributed to the final manuscript.

\section{Data}
The most up-to-date version of the models can be easily obtained and used thanks to the \texttt{pie-extended} Python package, available on \textit{Pypi} (\url{https://pypi.org/project/pie-extended/}), with the command \texttt{pie-extended download freem}. All the data are available online at \url{https://github.com/e-ditiones/LEM17}.

\begin{acks}
We would like to thank Lucence Ing and Frédéric Duval for 
their help regarding complicated annotation choices.
\end{acks}

\bibliographystyle{ACM-Reference-Format}
\bibliography{Lemmatisation}


\appendix
\section*{Appendix}

The full breakdown of the \textit{CornMol} and  \textit{FranText Open Access} corpora can be found in \citet{Camps2020}. We therefore describe only the \textit{Presto} data we have used.


\subsection*{Presto Gold}

\begin{tiny}
    \begin{tabularx}{\linewidth}{|XXlll|}
    \hline
        \textit{Author} &   \textit{Title} & \textit{1st ed.} & \textit{used ed.} & \textit{Tokens} \\
        SCÈVE M. & Saulsaye & 1547 & 1547 & 6,186 \\
        DU RYER P. & Lisandre et Caliste & 1632 & 1632 & 11,984  \\
        BUSSY-RABUTIN R. de & Les Lettres & 1666 & 1720 & 11,807 \\
        VOLTAIRE & Essay sur l’histoire générale et sur les moeurs & 1756 & 1756 & 16,224 \\
        RÉTIF DE LA BRETONNE N.-E. & Le Paysan perverti ou les Dangers de la ville & 1776 & 1776 & 16,141 \\
        \hline
        \end{tabularx}
\end{tiny}

    \subsection*{Presto core}
\begin{tiny}
    \begin{tabularx}{\linewidth}{|XXlll|}
    \hline
        \textit{Author} &   \textit{Title} & \textit{1st ed.} & \textit{used ed.} & \textit{Tokens} \\
        RABELAIS Fr. & Pantagruel & 1532 & 1542 & 45,371 \\
        RABELAIS Fr. & Gargantua & 1534 & 1542 & 50,499 \\
        FLORES J. de & La Deplourable fin de Flamete & 1535 & 1536 & 30,580 \\
        SCÈVE M. & Saulsaye & 1547 & 1547 & 6,182 \\
        DU BELLAY J. & La deffence, et illustration… & 1549 & 1549 & 20,974 \\
        ANONYME & Sottie pour le cry de la bazoche & 1549 & 1912 & 5,055 \\
        DES PÉRIERS B. & Nouvelles recreations et joyeux devis & 1558 & 1561 & 81,471 \\
        RONSARD P. de & Discours des Miseres de ce temps & 1562 & 1563 & 6,451 \\
        BUCHANAN G. & Jephté, ou le veu & 1567 & 1567 & 17,147 \\
        MONTAIGNE M. de & Essais & 1580 & 1580 & 224,183 \\
        URFÉ H. d’- & L’Astrée, première partie & 1607 & 1607 & 241,160 \\
        BÉROALDE DE VERVILLE Fr. & Le Moyen de parvenir & 1616 & 1616 & 148,460 \\
        COLLETET G. & Le Trébuchement de l’yvrongne & 1627 & 1627 & 2,713 \\
        BALZAC J.-L. GUEZ de & Le Prince & 1631 & 1631 & 61,629 \\
        DU RYER P. & Lisandre et Caliste & 1632 & 1632 & 21,507 \\
        PEIRESC N. Cl. & Lettres : t. 7 : lettres à divers : 1602-1637 & 1637 & 1898 & 87,931 \\
        ASSOUCY Ch. d’ & Poësies et lettres… & 1653 & 1653 & 32,459 \\
        BUSSY-RABUTIN R. de & Les Lettres & 1666 & 1720 & 192,622 \\
        SCUDÉRY M. de & Mathilde & 1667 & 1667 & 77,331 \\
        QUINAULT Ph. & Thésée & 1675 & 1675 & 13,329 \\
        ESPRIT J. & La Fausseté des vertus humaines & 1678 & 1710 & 175,131 \\
        CHARLEVAL J.-L. & Poésies & 1693 & 1759 & 12,111 \\
        REGNARD J.-Fr. & Les Folies amoureuses & 1704 & 1820 & 18,161 \\
        DACIER A. & Des causes de la corruption du goust & 1714 & 1714 & 65,273 \\
        RACINE L. & La grâce & 1720 & 1742 & 13,308 \\
        BOISSY L. de & Les Dehors trompeurs, ou l’Homme du jour & 1740 & 1813 & 24,796 \\
        VOLTAIRE & Essay sur l’histoire générale et sur les moeurs & 1756 & 1756 & 416,924 \\
        COLLECTIF & Encyclopédie (tome 7) & 1757 & 1757 & 1,600,920 \\
        ALEMBERT d’ & Lettre à M. Rousseau & 1759 & 1759 & 14,182 \\
        SAINT-LAMBERT J.-Fr. de & Les Saisons & 1769 & 1769 & 29,575 \\
        RÉTIF DE LA BRETONNE N.-E. & Le Paysan perverti ou les Dangers de la ville & 1776 & 1776 & 270,651 \\
        MIRABEAU H.-G. Riqueti & Lettres originales écrites du donjon… & 1780 & 1792 & 346,662 \\
        COLLIN D’HARLEVILLE J.-Fr. & L’Inconstant & 1786 & 1805 & 16,578 \\
        LA PÉROUSE J.-Fr. & Voyage de La Pérouse autour du monde & 1797 & 1797 & 338,344\\
    \hline
    \end{tabularx}
    
\end{tiny}


\end{document}